\documentclass{article}
\usepackage{microtype}
\usepackage{graphicx}
\usepackage{natbib}
\usepackage{amssymb,amsmath,amsthm}
\usepackage{subfigure}
\usepackage{booktabs} 
\newtheorem{definition}{Definition}
\newtheorem{theorem}{Theorem}
\DeclareMathOperator*{\argmin}{arg\, min} 
\usepackage{hyperref}
\usepackage{algorithm}
\usepackage[noend]{algcompatible}
\usepackage{fullpage}[cm]
\begin{document}

\title{Optimizing ROC Curves with a Sort-Based Surrogate Loss Function for Binary Classification and Changepoint Detection}
\author{Jonathan Hillman and \\
Toby Dylan Hocking --- toby.hocking@nau.edu}
\maketitle

\begin{abstract}
Receiver Operating Characteristic (ROC) curves are plots of true positive rate versus false positive rate which are useful for evaluating binary classification models, but difficult to use for learning since the Area Under the Curve (AUC) is non-convex.
ROC curves can also be used in other problems that have false positive and true positive rates such as changepoint detection.
We show that in this more general context, the ROC curve can have loops, points with highly sub-optimal error rates, and AUC greater than one.
This observation motivates a new optimization objective: rather than maximizing the AUC, we would like a monotonic ROC curve with AUC=1 that avoids points with large values for Min(FP,FN).
We propose a convex relaxation of this objective that results in a new surrogate loss function called the AUM, short for Area Under Min(FP, FN).
Whereas previous loss functions are based on summing over all labeled examples or pairs, the AUM requires a sort and a sum over the sequence of points on the ROC curve. 
We show that AUM directional derivatives can be efficiently computed and used in a gradient descent learning algorithm.
In our empirical study of supervised binary classification and changepoint detection problems, we show that our new AUM minimization learning algorithm results in improved AUC and comparable speed relative to previous baselines.
\end{abstract}

\section{Introduction}
\label{sec:introduction}
In supervised machine learning problems such as binary classification and changepoint detection, the goal is to learn a function for accurately predicting presence or absence of a class label.
In binary classification there is a prediction for each example; in changepoint detection there is a prediction (change or not) in between each data point in a sequence.
There are numerous ways to analyze prediction performance, but the simplest way is to calculate accuracy, which is the number or proportion of correctly classified labels.
However using accuracy as the evaluation metric can be problematic for data sets with imbalanced labels or for which the desired weighting of the labels is unknown \citep{cortes2004auc,Menon2013}.

A popular approach for comparing models in this context is by analyzing their Receiver Operating Characteristic (ROC) curves, which are plots of True Positive Rate (TPR) versus False Positive Rate (FPR) that have long been used in the signal processing literature \citep{egan1975signal}.
For data with $n$ labeled examples, most algorithms compute a predicted value $\hat y_i\in\mathbb R$ for each labeled example $i\in\{1,\dots,n\}$.
In binary classification the predicted value $\hat y_i$ is then compared to the threshold of zero to classify the example as either positive or negative.
True positives are examples $i$ with positive labels and positive predictions, whereas false positives are examples $i$ with negative labels and positive predictions.
A vector of predicted values $\mathbf{\hat{y}}=[\hat y_1\cdots\hat y_n]^\intercal\in\mathbb R^n$, one element for each labeled example $i$, can be mapped to a single (FPR,TPR) point in ROC space.
The different points on the ROC curve are obtained by adding a real-valued constant $c\in\mathbb R$ to each prediction in that vector, $\mathbf{\hat{y}}+c$.
Large constants $c$ result in FPR=TPR=1 and small constants result in FPR=TPR=0.
Therefore the Area Under the Curve (AUC) is an evaluation metric which accounts for all possible constants or possibilities for thresholding the prediction vector.
In binary classification, a perfect model has an AUC of 1, a constant model has an AUC of 0.5, and the minimum AUC is 0.

\begin{figure*}[ht]
\vskip 0.2in
\begin{center}
\includegraphics[width=0.55\textwidth]{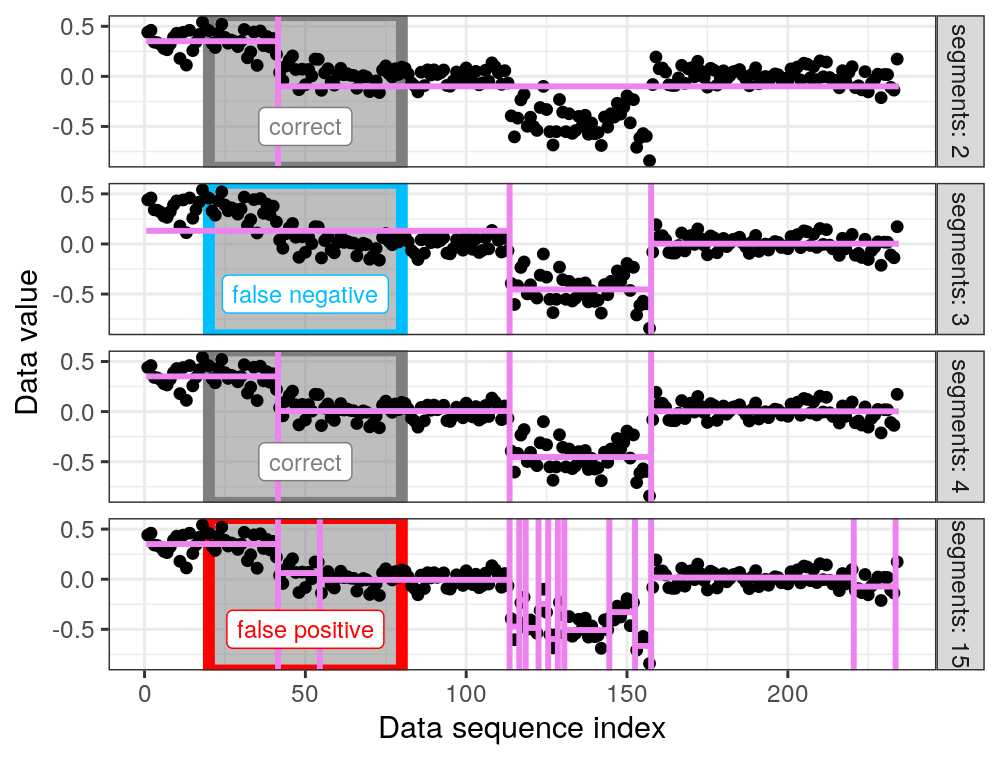}
\includegraphics[width=0.44\textwidth]{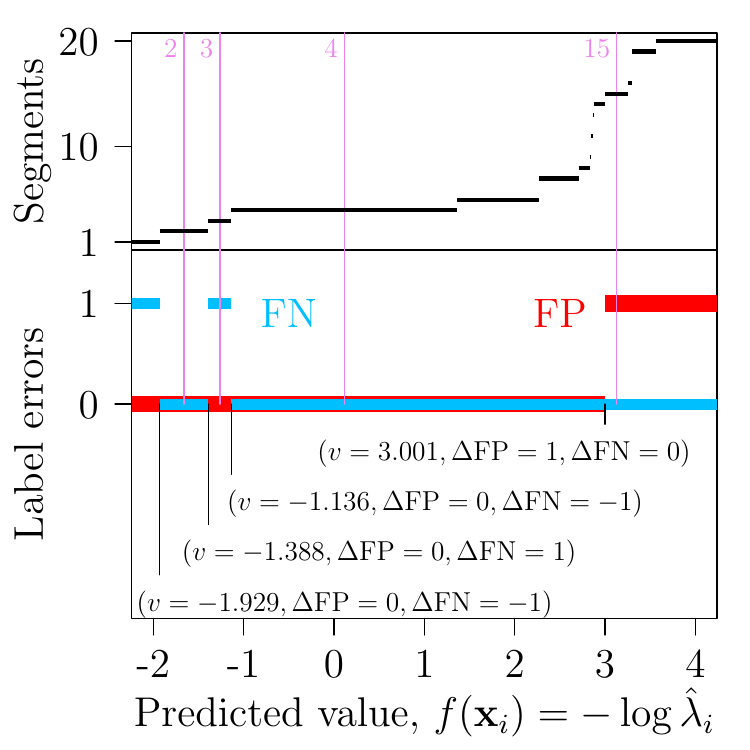}
\vskip -0.5cm
\caption{Example labeled changepoint detection data for which the FN function is non-monotonic. 
\textbf{Left:} a data sequence (black dots) with one label (grey rectangle) in which there should be exactly one predicted changepoint (false negative for no changes when segments=3, false positive for two changes when segments=15).  
\textbf{Right:} selected number of segments (top) and number of label errors (bottom) as a function of predicted values, with vertical violet lines indicating the models shown on the left, and vertical black lines for breakpoints in the error functions ($v$~for predicted value, $\Delta\text{FP},\Delta\text{FN}$ for changes in error functions at that value).
}
\label{fig:fn-not-monotonic}
\end{center}
\vskip -0.2in
\end{figure*}

When AUC is used as the evaluation metric in machine learning, the best algorithm is typically defined as the one that maximizes test AUC.
In such experiments, we would also like to use the AUC as the objective function for training the model (under the assumption that if the train and test sets are similar, maximizing train AUC should result in maximizing test AUC). 
However, since the AUC is a piecewise constant function of the predicted values, its gradient is zero almost everywhere, and it is therefore impossible to directly optimize using gradient descent algorithms.

In addition to studying binary classification, we also study supervised changepoint detection, in which we can also compute FPR and TPR.
For binary classification, we can compute a prediction vector which maximizes AUC in the same way as maximizing accuracy --- predict a positive value for each positive label, and a negative value for each negative label.
For changepoint detection, it is also trivial to compute a prediction vector that maximizes accuracy, but it can be non-trivial to compute a prediction vector that maximizes AUC.
The previous observations motivate this paper, which explores a new loss function and corresponding learning algorithm which we empirically show results in AUC maximization.

\subsection{Contributions and organization}

Our main contribution is the AUM, which is a new surrogate loss function defined as the Area Under the Minimum of false positive and false negative counts as a function of the prediction threshold. 
Whereas previously proposed loss functions for binary classification can be interpreted as convex surrogates of the zero-one loss (summed over examples) or Mann-Whitney statistic (summed over pairs of examples), our proposed AUM loss is an L1 relaxation of the total Min(FP,FN) (summed over points on the ROC curve, or distinct intervals of prediction thresholds).

In Section~\ref{sec:model} we give precise definitions of the AUC and our new AUM loss function.
In Section~\ref{sec:algorithms} we give an efficient algorithm for computing directional derivatives of the AUM with respect to predicted values, which we propose using in gradient descent learning algorithms.
In Section~\ref{sec:results} we provide an empirical study of supervised binary classification and changepoint detection problems, showing that minimizing the AUM corresponds to maximizing the AUC (with respect to train and test sets).
Section~\ref{sec:discussion} concludes with a discussion of the significance and novelty of our findings.

\subsection{Related work}
\label{sec:related-work}

\paragraph{Evaluating learned binary classification models.}
ROC curve analysis is a classical evaluation technique with origins in the signal processing literature \citep{egan1975signal}.
\citet{provost1997analysis} introduced a ROC convex hull method for evaluation.
\citet{ling2003auc} provide a proof of AUC consistency, whereas \citet{hand2009measuring} showed that it is incoherent with respect to classification costs, and proposed the ``H measure'' as an alternative.
\citet{martinez2017receiver} proposed a generalization to predicted values which have a non-monotone relationship with the desired class label.

\paragraph{Margin losses for binary classification.}
Margin losses can be interpreted as weighted convex surrogates of the zero-one loss, summed over all examples.
\citet{ferri2002learning, cortes2004auc} proposed expressions for the expected value
and the variance of the AUC for a fixed error rate, in an attempt to maximize the AUC with different nonlinear algorithms.
\citet{wang2015optimizing} incorporated unlabeled data to make an unsupervised AUC maximization algorithm.
Other algorithms have been proposed to obtain approximations of the global AUC value \citep{ rakotomamonjy2004optimizing, herschtal2004optimising, herschtal2006area, calders2007efficient}.
\citet{Han2010} proposed an active learning algorithm for computing a linear model that maximizes the AUC.
\citet{zhao2011online} implemented an online learning algorithm that maximizes the AUC.
\citet{Menon2013} analyzed statistical consistency under class imbalance, using a mean of true positive and true negative rates.
\citet{Scott2012} studied calibration for margin-based losses for binary classification.

\paragraph{Pairwise losses for binary classification.}
\citet{bamber1975area} is credited as the first to prove the equivalence of the ROC-AUC and the Mann-Whitney test statistic \citep{mann1947test}. 
This equivalence has led many authors to propose algorithms based on loss functions that are convex surrogates of the Mann-Whitney statistic.
\citet{yan2003optimizing} proposed a global approximation of AUC which was then used in several other algorithms.
For example, \citet{castro2008optimization} used that approxmation to propose the AUCtron algorithm which learns a linear model.
\citet{joachims2005multivariate} proposed a quadratic time support vector machine algorithm for AUC maximization based on a pairwise loss function, and
\citet{freund2003efficient} proposed a similar approach based on boosting.
\citet{narasimhan2013structural} extended this approach to the partial AUC, which is defined as the area under the curve between two false positive rates (not necessarily 0 and 1).
\citet{kotlowski2011bipartite} analyze how much risk and regret is increased for balanced margin-based losses, compared with pairwise losses.
\citet{rudin2005margin} showed that in some cases margin-based losses yield the same solution as a pairwise loss.
For stochastic optimization of the AUC, \citet{ying2016stochastic} proposed solving a saddle point problem involving a pairwise square loss, and 
\citet{yuan2020auc} proposed to extend this approach to the squared hinge loss.

\paragraph{Changepoint detection.} There are many algorithms for unsupervised changepoint detection \citep{aminikhanghahi2017survey,van2020evaluation}. 
In supervised changepoint detection, learning is often limited to grid search \citep{Hocking2013bioinformatics, Hocking2017bioinfo, Liehrmann2021chipseq}. 
More sophisticated learning algorithms use linear models and non-linear decision trees that minimize convex surrogates of the label error \citep{Hocking2013icml, Hocking2014,  Hocking2015, Drouin2017, Hocking2020psb}.
ROC curves and AUC are used to evaluate prediction accuracy of learned penalty functions \citep{Maidstone2016, Hocking2020jmlr}.


\begin{figure*}[t!]
\vskip 0.2in
\begin{center}
\includegraphics[height=2.2in]{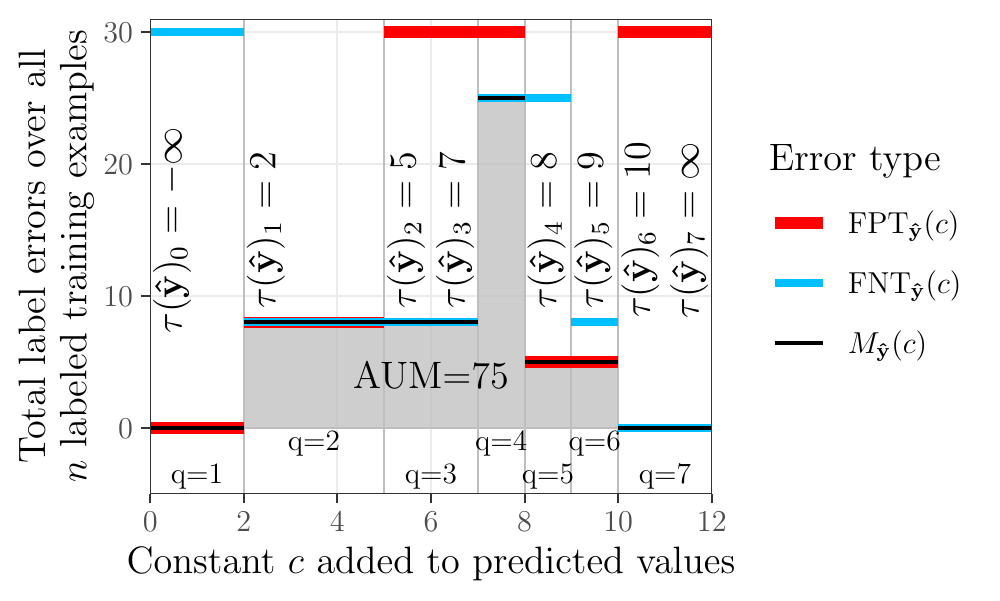}
\includegraphics[height=2.2in]{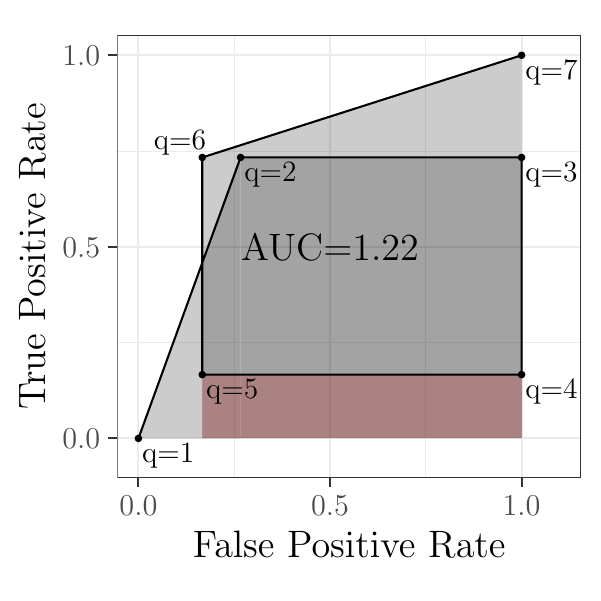}
\vskip -0.5cm
\caption{Synthetic example showing how non-monotonic total FP/FN functions (left) can result in a looping ROC curve (right).
The q values are indices in the sequence of points on the ROC curve (and corresponding intervals of the FP/FN functions).
\textbf{Left:} the AUM (grey area) is defined as the area under the minimum of total FP and FN functions, $M_{\mathbf{\hat{y}}}(t)=\min\{\text{FPT}_{\mathbf{\hat{y}}}(t),\text{FNT}_{\mathbf{\hat{y}}}(t)\}$.
For each interval $q\in\{1,\dots, 7\}$ of constant total error, the $\tau(\mathbf{\hat{y}})_q$ is the largest threshold in that interval.
\textbf{Right:} $\text{AUC}\geq 1$ due to the loop which results in double counting the dark grey area (but single counting the red area which is positive counted twice and negative counted once).
}
\label{fig:more}
\end{center}
\vskip -0.2in
\end{figure*}

\section{Models and Definitions}
\label{sec:model}
We begin by reviewing supervised binary classification and changepoint detection, then give definitions for AUC and the new AUM loss function.
\subsection{Review of supervised binary classification}

In supervised binary classification we are given a set of $n$ labeled training examples, $\{(\mathbf x_i, y_i)\}_{i=1}^n$ where $\mathbf x_i\in\mathbb R^p$ is an input feature vector for one example and $y_i\in\{-1,1\}$ is a binary output/label.
The goal of binary classification is to learn a function $f:\mathbb R^p\rightarrow \mathbb R$ which is used to compute real-valued predictions $\hat y_i=f(\mathbf x_i)$ with the same sign as the corresponding label $y_i$.
Margin-based learning algorithms such as logistic regression and support vector machines involve minimizing a convex surrogate loss function $\ell:\mathbb R\rightarrow \mathbb R$, summed over all training examples:
\begin{equation}
\label{eq:loss-sum-over-examples}
    \mathcal L(f) =  \sum_{i=1}^n \ell[ y_i f(\mathbf x_i) ].
\end{equation}
Large margin values $y_i f(\mathbf x_i)>0$ yield correct predictions, whereas small values $y_i f(\mathbf x_i)<0$ yield incorrect predictions.
Pairwise loss functions involve minimizing a convex loss summed over all pairs of positive and negative examples,
\begin{equation}
\label{eq:loss-sum-over-pairs}
    \mathcal L(f) =  
    \sum_{i:y_i = -1}\ \sum_{j: y_j = 1}
    \ell[ f(\mathbf x_j) - f(\mathbf x_i) ].
\end{equation}
Large pairwise difference values $f(\mathbf x_j) - f(\mathbf x_i) > 0$ yield correctly ranked pairs, whereas small pairwise difference values $f(\mathbf x_j) - f(\mathbf x_i) < 0$ yield incorrectly ranked pairs.
Typical choices for the convex surrogate loss function $\ell(z)$ include logistic $\log[1+\exp(-z)]$, linear hinge $(1-z)_+$, and squared hinge $(1-z)^2_+$, where $(\cdot)_+$ is the positive part function.
After a function $f$ has been learned using the training data, it can be evaluated by computing non-convex evaluation metrics such as the zero-one loss or the AUC with respect to a held-out test set.

\subsection{Review of supervised changepoint detection}

Here we give a brief overview of supervised changepoint detection; for details see \citep{Hocking2013icml}. 
We assume for each labeled training example $i\in\{1,\dots,n\}$ there is a corresponding data sequence vector $\mathbf z_i$ and label set $L_i$.
For example in Figure~\ref{fig:fn-not-monotonic} we show a data sequence $\mathbf z_i$ with a label set $L_i$ that contains one positive label (grey region in which there should be exactly one predicted changepoint).
Dynamic programming algorithms can be used on the data sequence $\mathbf z_i$ to compute a path of optimal changepoint models $\mathbf {\hat m}^k_i$ for different model sizes $k\in\{1,2,\dots\}$  \citep{Maidstone2016}.
For example in Figure~\ref{fig:fn-not-monotonic} (left) we show four models in the path (with $k=2,3,4,15$ segments).
The label set $L_i$ can be used to compute the number of false positive and false negative labels with respect to any predicted set of changepoints (false positives for too many changepoints, false negatives for not enough changepoints).
Each example $i$ also has a model selection function $k^*_i:\mathbb R^+_0 \rightarrow \{1,2,\dots\}$ which maps a non-negative penalty value $\hat \lambda_i$ to a selected model size $k^*_i(\hat \lambda_i)$ (Figure~\ref{fig:fn-not-monotonic}, right bottom).
We assume there is a fixed feature map $\phi$ which can be used to compute a feature vector $\mathbf x_i = \phi(\mathbf z_i)\in\mathbb R^p$ for each labeled example.

We want to learn a function $f:\mathbb R^p\rightarrow \mathbb R$ which inputs a feature vector and outputs a real-valued prediction that is used as a negative log penalty value, $f(\mathbf x_i) = -\log \hat \lambda_i$.
The goal is to predict model sizes $k^*_i(\hat \lambda_i)$ that result in minimal label errors. 
Since the label error function is non-convex like the zero-one loss in binary classification (Figure~\ref{fig:fn-not-monotonic}, right top), previous learning algorithms instead use the gradient with respect to a convex surrogate such as a hinge loss \citep{Hocking2013icml,Drouin2017} or a censored regression loss \citep{barnwal2021aftxgboost}.

\subsection{Definition of false positive and negative functions}
\label{sec:def-fp-fn}

In this paper, we assume the following general learning context in which supervised binary classification and changepoint detection are specific examples. 
For each labeled training example $i$, we have one or more labels such that there are at most $\text{FNP}_i\in\mathbb Z_+=\{0, 1, \dots\}$  false negatives possible and $\text{FPP}_i\in\mathbb Z_+$ false positives possible.
Given a real-valued prediction $\hat y_i=f(\mathbf x_i)\in\mathbb R$, we can use the labels to compute the number of predicted false positives $\text{FP}_i(\hat y_i)\in \{0, \dots, \text{FPP}_i\}$ and false negatives $\text{FN}_i(\hat y_i)\in\{0, \dots, \text{FNP}_i\}$.
The $\text{FP}_i,\text{FN}_i$ functions return the number of false positive and false negative labels for a given predicted value $\hat y_i$. 

\paragraph{Exact representation using breakpoints.} 
By convention we assume that the $\text{FP}_i,\text{FN}_i$ functions are piecewise constant and right continuous; 
that the $\text{FP}_i$ functions start at zero, $\lim_{x\rightarrow -\infty}\text{FP}_i(x)=0$; 
and the $\text{FN}_i$ functions end at zero, $\lim_{x\rightarrow \infty} \text{FN}_i(x)=0$. 
These assumptions ensure that (1) our proposed AUM loss function will always be finite, and (2) it can be computed efficiently by representing the error functions exactly using a finite set of breakpoints. For each breakpoint tuple $(v,\Delta\text{FP},\Delta\text{FN})$, $v\in\mathbb R$ is a predicted value threshold where there are changes $\Delta\text{FP},\Delta\text{FN}$ (discontinuity) in the error functions, $\text{FP}_i(v)-\lim_{x\rightarrow v^-} \text{FP}_i(x)= \Delta\text{FP}$ and similar for FN.
For example, in Figure~\ref{fig:fn-not-monotonic} the error functions can be exactly represented by a set of four such breakpoints.
These breakpoints will be used when computing our proposed loss function and learning algorithm.

\paragraph{Case of binary classification.} 
In the case of binary classification, for all positive examples $i:y_i=1$ we have 
 $\text{FPP}_i=0$, 
 $\text{FNP}_i=1$,
 $\text{FP}_i(\hat y) = 0$,
 $\text{FN}_i(\hat y) = I(\hat y < 0)$,
where $I$ is the indicator function (outputs 1 if argument is true, 0 otherwise).
For all negative examples $i:y_i=-1$,
  $\text{FPP}_i=1$,
  $\text{FNP}_i=0$,
  $\text{FP}_i(\hat y) = I(\hat y \geq 0)$, 
  $\text{FN}_i(\hat y) = 0$.
Note that $\text{FP}_i$ is either constant/zero (for positive examples) or non-decreasing (for negative examples), and $\text{FN}_i$ is either constant/zero (for negative examples) or non-increasing (for positive examples).
Since the prediction threshold is always zero in binary classification, these functions can be exactly represented by the breakpoint
$(v=0,\Delta\text{FP}=0,\Delta\text{FN}=-1)$ for all positive examples, and 
$(v=0,\Delta\text{FP}=1,\Delta\text{FN}=0)$ for all negative examples.

\paragraph{Case of changepoint detection.} 
In changepoint detection, we have more general $\text{FP}_i$ and $\text{FN}_i$ functions that can be non-monotonic, with arbitrary thresholds $v$ that can be computed in advance of learning $f$.
For example, in Figure~\ref{fig:fn-not-monotonic}, we show a data sequence with one positive label (in which there should be exactly one predicted changepoint).
Predicting no changepoint in this label results in a false negative, and predicting two changepoints in this label results in a false positive. 
Therefore, we have $\text{FPP}_i=\text{FNP}_i=1$ for this particular example $i$;
the false positive function is non-decreasing, $\text{FP}_i(\hat y_i) \approx I(\hat y_i \geq 3.001)$, and the false negative function is not monotonic, $\text{FN}_i(\hat y_i) \approx I[\hat y_i \in (\infty, -1.929)\cup (-1.388, -1.136)]$.
When the false negative function is non-monotonic, the true positive rate is also non-monotonic (the ROC curve can move down as well as up when prediction constant is increased).
Given a pre-computed path of changepoint models with loss/size values, the exact breakpoints $(v,\Delta\text{FP},\Delta\text{FN})$ in such error functions can be efficiently computed using a linear time algorithm \citep{Vargovich2020arxiv}.

\begin{figure*}[t]
\vskip 0.2in
\begin{center}
\includegraphics[height=4.5cm]{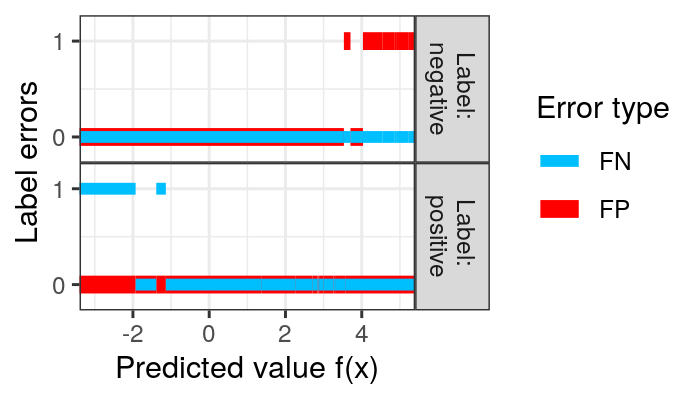}
\includegraphics[height=4.5cm]{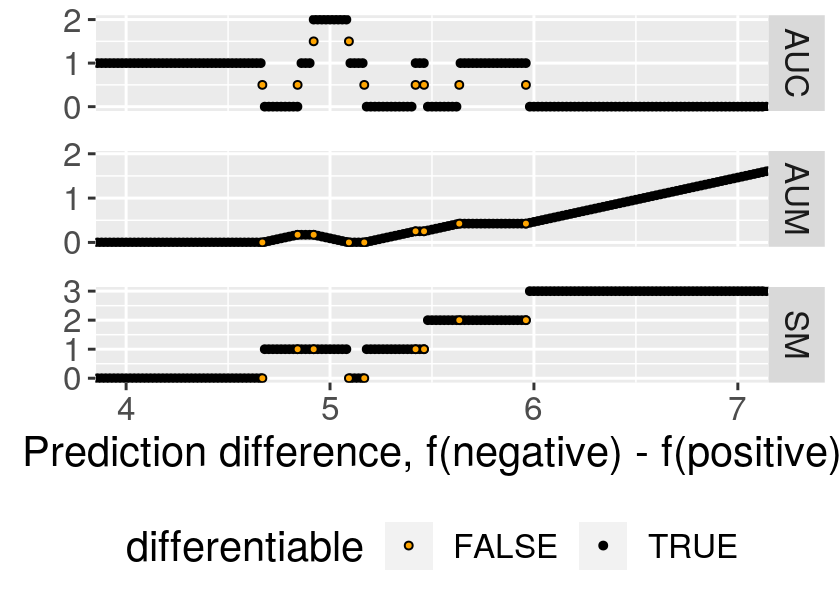}
\vskip -0.5cm
\caption{Real data example showing that it is possible to have AUC greater than one.
\textbf{Left:} error functions for two labeled examples in a real changepoint detection data set.
\textbf{Right:} AUC, AUM, and SM values as a function of the difference in predicted values between the two labeled examples.
When the difference is around 5, we have AUC=2.
The AUC and SM are piecewise constant functions of the predicted values, whereas the AUM is a continuous piecewise linear function.
At differentiable points, AUM=0 implies AUC=1 (but the converse is false).
}
\label{fig:aum-convexity}
\end{center}
\vskip -0.2in
\end{figure*}

\subsection{Definition of ROC curve and AUC}

In this section, we show how the previously defined functions can be used to define the ROC curve and the AUC.
Given a vector of predictions $\mathbf{\hat y}\in\mathbb R^n$, and a constant $c\in\mathbb R$ added to that vector, we define the False Positive Total (FPT) and False Negative Total (FNT) functions as
\begin{eqnarray}
 \text{FPT}_{\mathbf{\hat y}}(c) &=&
 \sum_{i=1}^n \text{FP}_i(\hat y_i + c),\label{eq:FPT} \\
 \text{FNT}_{\mathbf{\hat y}}(c) &=&
 \sum_{i=1}^n \text{FN}_i(\hat y_i + c).\label{eq:FNT}
\end{eqnarray}
The corresponding False Positive Rate (FPR) and True Positive Rate (TPR) functions are
\begin{eqnarray}
 \text{FPR}_{\mathbf {\hat y}}(c) &=&
 \frac{1}{\sum_{i=1}^n \text{FPP}_i } \label{eq:FPR}
 \text{FPT}_{\mathbf{\hat y}}(c), \\
 \text{TPR}_{\mathbf {\hat y}}(c) &=& \label{eq:TPR}
 1 - 
 \frac{1}{\sum_{i=1}^n \text{FNP}_i } 
 \text{FNT}_{\mathbf{\hat y}}(c).
\end{eqnarray}
The ROC curve for a given prediction vector $\mathbf {\hat y}$ is the plot of $\text{TPR}_{\mathbf {\hat y}}(c) $ versus $\text{FPR}_{\mathbf {\hat y}}(c) $ as the constant $c$ is varied from $-\infty$ to $\infty$.
The AUC is the integral,
\begin{equation}
    \text{AUC}(\mathbf{\hat y}) = 
    \int \text{TPR}_{\mathbf {\hat y}}(c) 
    d\, \text{FPR}_{\mathbf{\hat y}}(c).
\end{equation}
Assuming the $\text{FPT}_{\mathbf{\hat y}}(c)$ and $\text{FNT}_{\mathbf{\hat y}}(c)$ functions are piecewise constant (as in binary classification and changepoint detection) then the ROC curve can be described as a sequence of $Q$ points 
$\{(\text{fpr}
(\mathbf {\hat y})
_q, \text{tpr}
(\mathbf {\hat y})
_q)\}_{q=1}^Q$ in ROC space (note that this sequence of points depends on the predicted values $\mathbf {\hat y}$).
The first point $q=1$ corresponds to prediction threshold $t=-\infty$ which results in TPR=0 and FPR=0; the last point $q=Q$ is for $t=\infty$ which results in TPR=1 and FPR=1 (Figure~\ref{fig:more}).
The AUC can be computed using this sequence,
\begin{equation}
\label{eq:auc-seq}
\text{AUC}(\mathbf {\hat y}) = 
    \sum_{q=2}^{Q} 
    (\text{fpr}
    (\mathbf {\hat y})
    _{q} - \text{fpr}
    (\mathbf {\hat y})
    _{q-1})
    (\text{tpr}
    (\mathbf {\hat y})
    _{q-1} + \text{tpr}
    (\mathbf {\hat y})
    _{q})/2.
\end{equation}
In binary classification we have $\text{fpr}(\mathbf {\hat y})_{q-1} \leq \text{fpr}(\mathbf {\hat y})_{q}$ which means while tracing the ROC curve there are no moves to the left, and all the terms in the sum above are positive.
In fact this is also true of changepoint detection problems for which all the $\text{FP}_i$ functions are non-decreasing. 
In these cases the ROC curve is monotonic, with $\text{AUC}(\mathbf {\hat y})\in[0,1]$ for any predicted values $\mathbf {\hat y}$.

\subsection{Looping curves motivate optimizing ROC points instead of AUC}
If some of the $\text{FP}_i$ functions are not monotonic, then it is possible to have $\text{fpr}(\mathbf {\hat y})_{q-1} > \text{fpr}(\mathbf {\hat y})_{q}$ which results in negative terms in the AUC equation~(\ref{eq:auc-seq}). 
Furthermore if some of the $\text{FN}_i$ functions are not monotonic, then the ROC curve can contain loops which double count some area in ROC space.
For a simple synthetic example, consider the FP/FN functions and corresponding ROC curves in Figure~\ref{fig:more}.
In this example, the ROC curve has area $\approx 1.22$ because it contains a loop which double-counts a large portion of ROC space.
This looping implies existence of ROC points $q$ which are highly sub-optimal in terms of TPR and/or FPR (for example, $q\in\{2,3,4,5\}$ in Figure~\ref{fig:more}).
By ``sub-optimal'' we mean ROC points that occur more toward the bottom right of ROC space, where both FP and FN are large, as is $\min\{\text{FP},\text{FN}\}$.
Thus we observe an association between looping ROC curves, large AUC, and existence of some thresholds with highly sub-optimal error rates.

Although this seems like a rare phenomenon, we have seen this occur in real data. For example, in Figure~\ref{fig:fn-not-monotonic} it is clearly possible to have error functions which are non-monotonic.
Furthermore, in Figure~\ref{fig:aum-convexity} (left) we show error curves for two labeled examples from another real changepoint detection data set.
One example has a positive label that results in a non-monotonic false negative function, and the other example has a negative label that results in a non-monotonic false positive function.
When the predicted value for the negative example is about 5 greater than the predicted value for the positive example, we observe AUC=2 (Figure~\ref{fig:aum-convexity}, right).
This is the result of a loop in ROC space, indicating that there is some prediction threshold that results in (FPR=1,TPR=0) which corresponds to 100\% error and 0\% accuracy.
Therefore, maximizing the AUC may not be a desirable optimization objective, because it can result in ROC curves with loops and points in the lower right, with large $\min\{\text{FP},\text{FN}\}$ values. 
Instead, we propose as an optimization objective to minimize the total $\min\{\text{FP},\text{FN}\}$ over all ROC points.
This objective means that ideal ROC points would be moved away from the lower right, toward the upper left, with the best case being a monotonic ROC curve with AUC=1.
We formalize this idea in the next section, and provide a new surrogate loss function for this objective.

\subsection{Proposed surrogate loss function (AUM)}
In this section we propose the AUM loss function, which is short for Area Under Min(FP,FN).
The intuition behind the AUM is that we want to minimize the number of prediction thresholds that result in large error rates.
To formally define the AUM we must first define the minimum of total false positives and false negatives, 
\begin{equation}
\label{eq:M}
    M_{\mathbf{\hat y}}(c) = 
    \min\{
    \text{FPT}_{\mathbf{\hat y}}(c),
    \text{FNT}_{\mathbf{\hat y}}(c)
    \}.
\end{equation}
Then we define the AUM as 
\begin{equation}
\label{eq:AUM}
    \text{AUM}(\mathbf {\hat y}) =
    \int_{-\infty}^{\infty}
    M_{\mathbf {\hat y}}(c)\, dc.
\end{equation}
In summary, we compute AUM by integrating the $M$ function over all possible constants $c\in\mathbb R$.
Geometrically, this corresponds to the area under the minimum of total false positive and false negative functions (Figure~\ref{fig:more}, left).

\subsection{Interpretation as L1 relaxation of total Min(FP,FN) over ROC points}
\label{sec:algo-aum-overview}
Computing the AUM is similar to the AUC, in that we must compute error/accuracy rates for each possible prediction threshold. 
First let $\{(
\text{fpt}
(\mathbf {\hat y})
_q, \text{fnt}
(\mathbf {\hat y})
_q,
 \tau
(\mathbf {\hat y})
_q
)\}_{q=1}^Q$ 
be a sequence of $Q$ tuples, each of which corresponds to a point on the ROC curve (Figure~\ref{fig:more}, right).
The fpt/fnt are false positive/negative totals whereas $\tau$ are values such there is a change/threshold at $M_{\mathbf{\hat y}}(\tau)$. 
As shown in Figure~\ref{fig:more} we assume these values are increasing, $ -\infty = \tau
(\mathbf {\hat y})
_0 < \cdots <  \tau
(\mathbf {\hat y})
_Q = \infty$.
For each $q\in\{1,\dots,Q\}$ there is a corresponding interval of values $c$ between $\tau(\mathbf {\hat y})_{q-1}$ and $\tau(\mathbf {\hat y})_q$
such that 
$\text{FPT}_{\mathbf{\hat y}}(c)=\text{fpt}(\mathbf {\hat y})_q+t$
and
$\text{FNT}_{\mathbf{\hat y}}(c)=\text{fnt}(\mathbf {\hat y})_q+t$
for all $c\in(\tau(\mathbf {\hat y})_{q-1}, \tau(\mathbf {\hat y})_q)$
(Figure~\ref{fig:more}, left).
Then we define $m(\mathbf {\hat y})_q = \min\{
    \text{fpt}(\mathbf {\hat y})_q , \, 
    \text{fnt}(\mathbf {\hat y})_q
\}$ and so since 
$m(\mathbf {\hat y})_1=m(\mathbf {\hat y})_Q=0$ the area under those intervals is zero, and the AUM can be computed by summing over all of the other intervals,
\begin{equation}
\label{eq:AUM-computation}
    \text{AUM}(\mathbf {\hat y}) =
    \sum_{q=2}^{Q-1}
    [ \tau(\mathbf {\hat y})_{q} - \tau(\mathbf {\hat y})_{q-1} ]
    m(\mathbf {\hat y})_q.
\end{equation}
For example, in Figure~\ref{fig:more} there are $Q=7$ tuples (points on the ROC curve), five of which result in a positive AUM value, resulting in a total AUM is 75.

The AUM can be interpreted as an L1 relaxation of the following non-convex \textbf{S}um of \textbf{M}in(FP,FN) function, 
\begin{equation}
\label{eq:SM-computation}
    \text{SM}(\mathbf {\hat y}) =
    \sum_{q=2}^{Q-1}
    I[ \tau(\mathbf {\hat y})_{q} \neq \tau(\mathbf {\hat y})_{q-1} ]
    m(\mathbf {\hat y})_q =
    \sum_{q:\tau(\mathbf {\hat y})_{q} \neq \tau(\mathbf {\hat y})_{q-1} }
    m(\mathbf {\hat y})_q.
\end{equation}
The difference is that the L1 norm of $\tau(\mathbf {\hat y})_{q} - \tau(\mathbf {\hat y})_{q-1}$ in (\ref{eq:AUM-computation}) has been changed to the L0 pseudo-norm in (\ref{eq:SM-computation}). 
The indicator function $I$ takes values in zero and one, which means the SM is the sum of $m(\mathbf{\hat y})_q$ values over all distinct ROC points $q$. 
Geometrically, the SM function measures total distance of distinct points on the ROC curve from the left (FPR=0) or top (TPR=1, FNR=0) of ROC space.
A visualization of the piecewise constant SM loss as a function of the predicted values $\hat y_i$ shows that the AUM is indeed a continuous relaxation that is differentiable almost everywhere (Figure~\ref{fig:aum-convexity}, right).

\subsection{Properties of AUM}
A few interesting properties of the AUM should be immediately apparent.
\begin{description}
    \item[Finite.] 
Since $\text{FP}_i(-\infty)=0$ and $\text{FN}_i(\infty)=0$ for all $i$ (by assumption), we have $M(\mathbf{\hat y}\pm\infty)=0$, which means 
the AUM integral~(\ref{eq:AUM}) is finite, $\text{AUM}(\mathbf{\hat y})<\infty$.
\item[Non-negative.] Since the label error is never negative, the AUM integral is never negative, $\text{AUM}(\mathbf{\hat y})\geq 0$.
\item[Convexity.] If all of the $\text{FP}_i$ and $\text{FN}_i$ functions are monotonic, then the AUM is convex.
\item[Other properties.] The AUM is continuous, piecewise linear, and differentiable almost everywhere. 
\end{description}
It can be seen that non-differentiable points in the AUM coincide with discontinuities in AUC (Figure~\ref{fig:aum-convexity}, right).
Furthermore
at differentiable points, 
if $\text{AUM}(\mathbf{\hat y})=0$ then
$\text{AUC}(\mathbf{\hat y})=1$; 
the converse is clearly false, however (Figure~\ref{fig:aum-convexity}, right).
In this figure it can also be seen that for all predictions with maximal AUC=2, we have $\text{AUM}>0$, which means there are some prediction thresholds which result in FP=FN=1.
This observation suggests that instead of maximizing AUC, it may be preferable to minimize AUM as a surrogate loss, to obtain AUC=1. 
We therefore propose an AUM gradient descent learning algorithm in the next section.

\section{Algorithms}
\label{sec:algorithms}

In this section we propose algorithms for computing the AUM and its gradient, and for learning a linear model.

\subsection{Details of AUM computation}
\label{sec:algo-aum-details}

In previous sections we have assumed that a sequence of $Q$ points on the ROC curve (or equivalently intervals of FP/FN functions) can be computed, and in this section we explain how to do that.
Recall from Section~\ref{sec:def-fp-fn} that the $\text{FP}_i,\text{FN}_i$ functions have an exact representation in terms of breakpoints.
Let there be a total of $B$ breakpoints in the error functions over all $n$ labeled training examples, where each breakpoint $b\in\{1,\dots, B\}$ is represented by the tuple $(v_b, \Delta\text{FP}_b, \Delta\text{FN}_b, \mathcal I_b)$.
The $\mathcal I_b\in\{1,\dots,n\}$ is an example index, so there are changes $\Delta\text{FP}_b, \Delta\text{FN}_b$ at predicted value $v_b\in\mathbb R$ in the error functions $\text{FP}_{\mathcal I_b},\text{FN}_{\mathcal I_b}$ (Figure~\ref{fig:fn-not-monotonic}). 
For example in binary classification, there are $B=n$ breakpoints (same as the number of labeled training examples); for each breakpoint $b\in\{1,\dots,B\}$ we have $v_b=0$ and $\mathcal I_b=b$. 
For breakpoints $b$ with positive labels $y_b=1$ we have $\Delta\text{FP}=0,\Delta\text{FN}=-1$,
and for negative labels $y_b=-1$ we have $\Delta\text{FP}=1,\Delta\text{FN}=0$.
In changepoint detection we have more general error functions, which may have more than one breakpoint per example.
For example the labeled data sequence shown in Figure~\ref{fig:fn-not-monotonic} is a single labeled training example $i$ with error functions that can be represented by four breakpoints $b$ with the same $\mathcal I_b=i$ value.

Given a prediction vector $\mathbf{\hat y}=[\hat y_1 \cdots \hat y_n]^\intercal\in\mathbb R^n$ we can compute a prediction threshold $t_b= v_b - \hat y_{\mathcal I_b}$ for each breakpoint $b\in\{1,\dots,B\}$.
The prediction thresholds $t\in\{t_1,\dots,t_B\}$ are where there are changes in the total error functions $\text{FPT}_{\mathbf{\hat y}}(t),\text{FNT}_{\mathbf{\hat y}}(t)$ defined in equations (\ref{eq:FPT}--\ref{eq:FNT}).
These functions can be exactly represented by the sequences of $b\in\{1,\dots,B\}$ error values
\begin{eqnarray}
  \underline{\text{FP}}_b &=& \sum_{j: t_j < t_b} \Delta\text{FP}_j,\label{eq:first-fp-fn-over-under} \\
  \overline{\text{FP}}_b &=& \sum_{j: t_j \leq t_b} \Delta\text{FP}_j, \\
  \underline{\text{FN}}_b &=& \sum_{j: t_j \geq t_b} - \Delta\text{FN}_j, \\
  \overline{\text{FN}}_b &=& \sum_{j: t_j > t_b} - \Delta\text{FN}_j.\label{eq:last-fp-fn-over-under}
\end{eqnarray}
The $\underline{\text{FP}}_b, \underline{\text{FN}}_b$ are the error values before the threshold $t_b$, whereas $\overline{\text{FP}}_b, \overline{\text{FN}}_b$ are the error values after the threshold. 
We sort the breakpoints by threshold, yielding a permutation $\{s_1,\dots, s_B\}$ of the indices $\{1,\dots,B\}$ such that for every $q\in\{2,\dots,B\}$ we have $t_{s_{q-1}} \leq  t_{s_q}$.
All of the error values $\underline{\text{FP}}_b,\overline{\text{FP}}_b,\underline{\text{FN}}_b,\overline{\text{FN}}_b$, 
for every $b\in\{1,\dots,B\}$, can then be computed via a modified cumulative sum (starting with 
$\underline{\text{FP}}_1=0$ and
$\overline{\text{FN}}_B=0$).
In terms of the notation of Section~\ref{sec:algo-aum-overview}, we have $Q=B+1$ points (not necessarily unique) on the ROC curve such that $\tau(\mathbf {\hat y})_{q} = t_{s_q}$ and $m(\mathbf {\hat y})_q = \min\{
    \underline{\text{FP}}_{s_q}, 
    \underline{\text{FN}}_{s_q}
    \} = \min\{
    \overline{\text{FP}}_{s_{q-1}}, 
    \overline{\text{FN}}_{s_{q-1}}
    \}$.
The AUM can then be computed via 
\begin{eqnarray}
    \text{AUM}(\mathbf{\hat y}) &=& 
    \sum_{q=2}^B 
    (t_{s_q} - t_{s_{q-1}}) 
    \min\{
    \overline{\text{FP}}_{s_{q-1}}, 
    \overline{\text{FN}}_{s_{q-1}}\label{eq:min_above}
    \},\\
    &=& 
    \sum_{q=2}^B 
    (t_{s_q} - t_{s_{q-1}}) 
    \min\{
    \underline{\text{FP}}_{s_q}, 
    \underline{\text{FN}}_{s_q}\label{eq:min_below}
    \}. 
\end{eqnarray}
The equations above state that the AUM can be computed by multiplying each threshold difference $t_{s_q} - t_{s_{q-1}}$ by the minimum below breakpoint $s_q$ (\ref{eq:min_below}) or above breakpoint $s_{q-1}$ (\ref{eq:min_above}). 
Since the slowest operation is the sort, the overall time complexity for computing the AUM is $O(B\log B)$.

\begin{algorithm}[t]
  \caption{\label{alg:gradient-computation}AUM and Directional Derivatives}
\begin{algorithmic}[1]
  \STATE {\bfseries Input:} 
  Predictions $\mathbf{\hat y}\in\mathbb R^n$, 
  breakpoints in error functions $v_b,\Delta\text{FP}_b,\Delta\text{FN}_b,\mathcal I_b$ for all $b\in\{1,\dots,B\}$.
  \STATE Initialize to zero the $\text{AUM}\in\mathbb R$ and directional derivative matrix $\mathbf D\in\mathbb R^{n\times 2}$.\label{line:init-zero}
  \STATE $t_b\gets v_b - \hat y_{\mathcal I_b}$ for all $b$.\label{line:compute-thresh}
  \STATE $s_1,\dots,s_B\gets \textsc{SortedIndices}(t_1,\dots,t_B).$\label{line:sorted-indices}
  \STATE Compute $\underline{\text{FP}}_b,\overline{\text{FP}}_b,\underline{\text{FN}}_b,\overline{\text{FN}}_b$ for all $b$ using $s_1,\dots,s_B$ and\label{line:fp-fn-sum-over-under} (\ref{eq:first-fp-fn-over-under}--\ref{eq:last-fp-fn-over-under}).
  \FOR{$b\in\{2,\dots,B\}$}\label{line:for-intervals}
  \STATE $\text{AUM} \text{ += } (t_{s_b} - t_{s_{b-1}}) \min\{\underline{\text{FP}}_b, \overline{\text{FN}}_b\} $.\label{line:AUM}
  \ENDFOR
  \FOR{$b\in\{1,\dots,B\}$}\label{line:for-breakpoints}
  \STATE\label{line:D_lo} $D_{\mathcal I_b,1} \text{ += } \min\{
  \overline{\text{FP}}_b , 
  \overline{\text{FN}}_b 
  \}
  -
  \min\{
  \overline{\text{FP}}_b - \Delta\text{FP}_b, 
  \overline{\text{FN}}_b - \Delta\text{FN}_b
  \}$
  \STATE\label{line:D_hi} $D_{\mathcal I_b,2} \text{ += } \min\{
  \underline{\text{FP}}_b + \Delta\text{FP}_b, 
  \underline{\text{FN}}_b + \Delta\text{FN}_b
  \}
  -
  \min\{
  \underline{\text{FP}}_b , 
  \underline{\text{FN}}_b 
  \}$
  \ENDFOR
  \STATE {\bfseries Output:} AUM and matrix $\mathbf D$ of directional derivatives.
\end{algorithmic}
\end{algorithm}

\subsection{Gradient computation}
\label{sec:gradient-computation}

First we note that the AUM function is not differentiable everywhere, so the gradient is not defined everywhere (e.g., orange dots in Figure~\ref{fig:aum-convexity}, right).
Second we note that the AUM function can be non-convex, in which case the sub-differential from convex analysis is not defined \citep{rockafellar-1970a}.
Instead, we propose an algorithm for computing the AUM directional derivatives, which are defined everywhere. 
We recall the general definition of a directional derivative.

\begin{definition}
Given vectors $\mathbf x,\mathbf v\in\mathbb R^n$ and a function $f:\mathbb R^n\rightarrow \mathbb R$, the directional derivative of $f$ at $\mathbf x$ in the direction of $\mathbf v$ is the function $\nabla_{\mathbf v} f: \mathbb R^n \rightarrow \mathbb R$ given by
\begin{equation}
\label{eq:directional-derivative}
    \nabla_{\mathbf v} f(\mathbf x) = 
    \lim_{h\rightarrow 0}
    \frac{f(\mathbf x + h\mathbf v) - 
    f(\mathbf x)}{h}.
\end{equation}
\end{definition}
We would like to compute $\nabla_{\mathbf v}\text{AUM}(\mathbf{\hat y})$, for a set of directions $\mathbf v$.
We are interested in computing the directional derivative along a single dimension $i\in\{1,\dots,n\}$, in either the negative or positive direction, which correspond to using direction vectors $\mathbf v$ with $-1$ or 1 at the $i$-th position, and zeros at each other position, 
\begin{eqnarray}
\mathbf v(-1, i) &=& \left[\begin{array}{ccccc}
0 & \cdots & -1 & \cdots & 0
\end{array}\right]^\intercal,\\
\mathbf v(1, i) &=& \left[\begin{array}{ccccc}
0 & \cdots & 1 & \cdots & 0
\end{array}\right]^\intercal.
\end{eqnarray}
The intuition of these direction vectors is that each will give us the rate of change of AUM, if a single prediction $i$ is either increased or decreased.
We propose an algorithm for efficiently computing the following $n\times 2$ matrix of directional derivatives,
\begin{equation}
\mathbf D_f(\mathbf x) = 
    \left[\begin{array}{ccc}
\nabla_{\mathbf v(-1,1)} f(\mathbf x) &
\cdots &
\nabla_{\mathbf v(-1,n)} f(\mathbf x) \\
\nabla_{\mathbf v(1,1)} f(\mathbf x) &
\cdots &
\nabla_{\mathbf v(1,n)} f(\mathbf x) 
    \end{array}\right]^\intercal.
\end{equation}
We will compute $\mathbf D_\text{AUM}(\mathbf {\hat y})$, which is the matrix of directional derivatives for a given prediction vector.
If we have equality of elements of all rows of this matrix, i.e., $
\nabla_{\mathbf v(-1,i)} \text{AUM}(\mathbf {\hat y})
=
\nabla_{\mathbf v(1,i)} \text{AUM}(\mathbf {\hat y})$ for all $i$, then the gradient does exist at $\mathbf {\hat y}$ and is equal to that value.
The following theorem shows how to compute the elements of this matrix.
\begin{theorem}
\label{thm:directional-derivs}
The AUM directional derivatives for a particular example $i\in\{1,\dots,n\}$ can be computed using the following equations.
\end{theorem}
\begin{eqnarray}
  \nabla_{\mathbf v(-1,i)} \text{AUM}(\mathbf{\hat y}) &=& 
  \sum_{b: \mathcal I_b = i}
  \min\{
  \overline{\text{FP}}_b , 
  \overline{\text{FN}}_b 
  \}
  -
  \min\{
  \overline{\text{FP}}_b - \Delta\text{FP}_b, 
  \overline{\text{FN}}_b - \Delta\text{FN}_b
  \},\\
  \nabla_{\mathbf v(1,i)} \text{AUM}(\mathbf{\hat y}) &=& 
  \sum_{b: \mathcal I_b = i}
  \min\{
  \underline{\text{FP}}_b + \Delta\text{FP}_b, 
  \underline{\text{FN}}_b + \Delta\text{FN}_b
  \}
  -
  \min\{
  \underline{\text{FP}}_b , 
  \underline{\text{FN}}_b 
  \}.
\end{eqnarray}
\begin{proof}
We can compute AUM using either (\ref{eq:min_above}) or (\ref{eq:min_below}).
To compute a directional derivative we need to evaluate $\text{AUM}[\mathbf{\hat y} + h \mathbf v(d,i)]$.
To do that we must use (\ref{eq:min_above}) if $d= -1$ and (\ref{eq:min_below}) if $d=1$. 
First consider the case of $d=1$, if $t_{s_{q-1}} = t_{s_q}$, and $\mathcal I_{s_{q-1}}= i$.
In that case the analogous term in $\text{AUM}[\mathbf{\hat y} + h \mathbf v(d,i)]$ will have $v_{s_{q-1}} - \hat y_{s_{q-1}}-h < v_{s_q}-\hat y_{s_q}$.
Then the min in the corresponding term must be $\min\{\underline{\text{FP}}_{s_{b-1}}+\Delta\text{FP}_{b-1},\underline{\text{FN}}_{s_{b-1}}+\Delta\text{FP}_{b-1}\}$ and so
\begin{eqnarray}
\nabla_{\mathbf v(1,i)} \text{AUM}(\mathbf{\hat y})
&=& \sum_{b=2}^{B} \left( I[\mathcal I_{s_{b-1}} = i] - I[\mathcal I_{s_b}=i]\right) \min\{\underline{\text{FP}}_{s_{b-1}}+\Delta\text{FP}_{s_{b-1}},
\underline{\text{FN}}_{s_{b-1}}+\Delta\text{FP}_{s_{b-1}}
\}, \\
&=& \sum_{b=2}^B I[\mathcal I_{s_{b-1}} = i]  \min\{\underline{\text{FP}}_{s_{b-1}}+\Delta\text{FP}_{s_{b-1}},\underline{\text{FN}}_{s_{b-1}}+\Delta\text{FP}_{s_{b-1}}\}\nonumber\\
&&
-\sum_{b=2}^B I[\mathcal I_{s_b}=i] \min\{\underline{\text{FP}}_{s_{b-1}}+\Delta\text{FP}_{s_{b-1}},\underline{\text{FN}}_{s_{b-1}}+\Delta\text{FP}_{s_{b-1}}\}, \\
&=& \sum_{b=1}^B I[\mathcal I_{s_{b}} = i] \min\{\underline{\text{FP}}_{s_{b}}+\Delta\text{FP}_{s_{b}},\underline{\text{FN}}_{s_{b}}+\Delta\text{FP}_{s_{b}}\} \nonumber\\
&&-\sum_{b=1}^B I[\mathcal I_{s_b}=i] \min\{\underline{\text{FP}}_{s_{b}},\underline{\text{FN}}_{s_{b}}\}.
\end{eqnarray}
The first equality comes from the definition of the directional derivative, the second distributes the min, and the third re-writes some of the $b-1$ indices as $b$.
The sum can be extended to start at $b=1$ since $\underline{\text{FP}}_{s_{1}}=0$ (the first min is zero). 
Finally, re-writing the sums and removing the indicator functions obtains the desired result.
The proof for the case of $d= -1 $ is analogous.
\end{proof}

\begin{figure*}[t]
\vskip 0.2in
\begin{center}
\includegraphics[height=5.5cm]{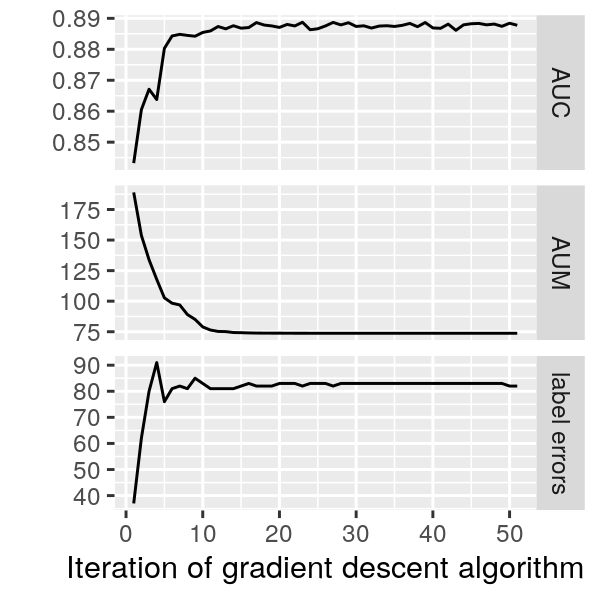}
\includegraphics[height=5.5cm]{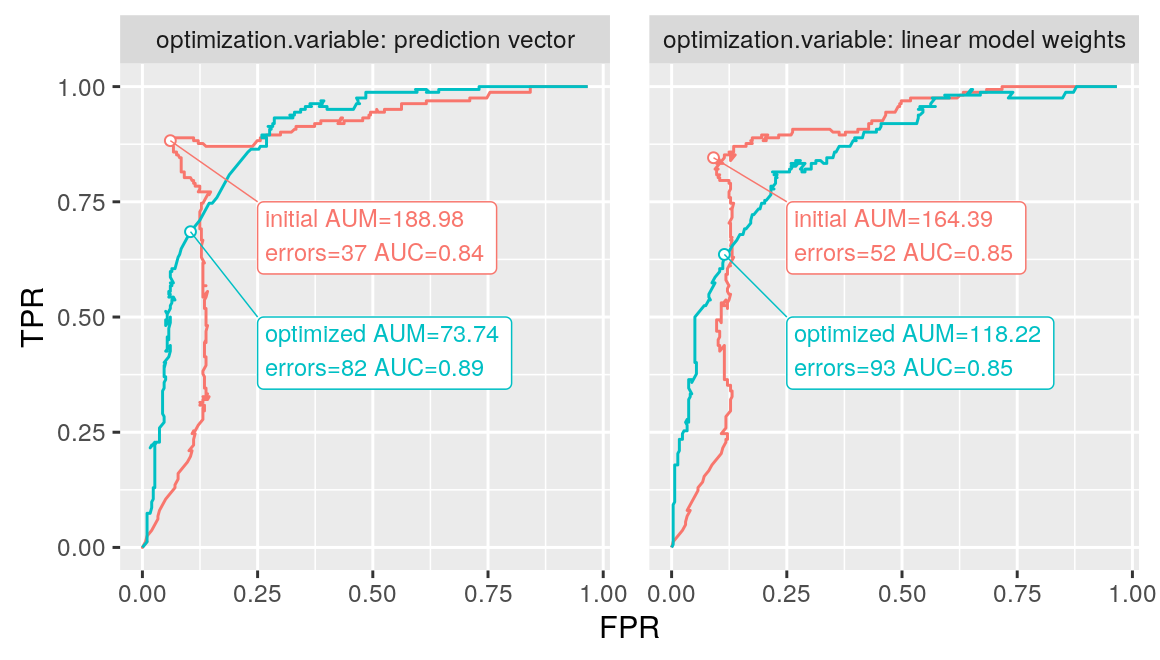}
\vskip -0.5cm
\caption{In a real changepoint detection problem with $n=54$ labeled examples, minimizing the AUM tends to increase train AUC and label error rates.
\textbf{Left:} we used AUM gradient descent on the $n$-vector of predicted values $\mathbf{\hat y}$ in the train set (initial predictions chosen by minimizing label errors for each example). Decreases in AUM happen during the same iterations as increases in AUC and label error rates. 
\textbf{Middle:} ROC curves before and after using AUM gradient descent on the $n$-vector of predicted values (dot shows a point on each ROC curve with minimum label errors).
\textbf{Right:} optimizing the $p$-vector of weights in a linear model ($p=27$ features, initial weights minimize an un-regularized squared hinge loss summed over labeled examples). 
Note that AUC=0.85 is unchanged by the optimization but AUM decreases and error rate increases.
}
\label{fig:aum-optimized}
\end{center}
\vskip -0.2in
\end{figure*}


\subsection{Pseudocode and complexity analysis}
We propose to compute the matrix of directional derivatives using Algorithm~\ref{alg:gradient-computation} which inputs a prediction vector $\mathbf {\hat y}\in\mathbb R^n$ and an exact description of the error functions in terms of breakpoints $v_b,\Delta\text{FP}_b,\Delta\text{FN}_b,\mathcal I_b$. 
The first step is to compute thresholds $t_b=v_b - \hat y_{\mathcal I_b}$ in the total error functions (line~\ref{line:compute-thresh}) which are then sorted (line~\ref{line:sorted-indices}).
The sorted indices are then used to compute modified cumulative sums of false positives and false negatives, $\underline{\text{FP}}_b,\overline{\text{FP}}_b,\underline{\text{FN}}_b,\overline{\text{FN}}_b$, for each breakpoint $b$ (line~\ref{line:fp-fn-sum-over-under}).
Note that for FP we start with $\underline{\text{FP}}_1=0$ and add in the forward direction, whereas for FN we start with $\overline{\text{FN}}_B=0$ and subtract in the reverse direction.
Each iteration of the for loop over intervals of threshold values (line~\ref{line:for-intervals}) adds to the AUM (line~\ref{line:AUM}).
Each iteration of the for loop over breakpoints $b$ (line~\ref{line:for-breakpoints}) adds to row $\mathcal I_b$ of the directional derivative matrix.
The first column stores the derivative in the negative direction (line~\ref{line:D_lo}) and the second column stores the derivative in the positive direction (line~\ref{line:D_hi}).
The time complexity of Algorithm~\ref{alg:gradient-computation} is log-linear $O(B\log B)$ because of the sort (line~\ref{line:sorted-indices}).
Since there is at least one breakpoint in each of the $n$ example-specific error functions, the total number of breakpoints $B\geq n$.
Therefore, computing the AUM and its directional derivatives is $O(n\log n)$, asymptotically slower than $O(n)$ margin-based loss functions by a log factor.

\subsection{Gradient descent algorithm for predicted values}
\label{sec:gradient-descent}

We propose gradient descent optimization algorithms that use the AUM directional derivatives computed using Theorem~\ref{thm:directional-derivs}
and Algorithm~\ref{alg:gradient-computation}.
First to study how minimizing the train AUM affects the train AUC, we propose to optimize the $n$-vector of predictions $\mathbf{\hat{y}}$.
When the AUM is non-convex, the initialization of the algorithm is important. Therefore when optimizing the vector of predictions, we propose initializing each $\hat y_i$ to a value with minimum label errors, 
\begin{equation}
    \hat y_i^{(0)} = \argmin_x 
    \text{FP}_i(x) + 
    \text{FN}_i(x).
\end{equation}
After initialization we need to compute a descent direction; typically the negative gradient is used.
Recall that since the AUM has non-differentiable points, the gradient is not defined at these points.
The ``gradient'' we propose to use is the mean of the two columns of the directional derivative matrix, $\mathbf{\bar D}_{\text{AUM}}(\mathbf{\hat y})\in\mathbb R^n$, with each element $i\in\{1,\dots,n\}$ of this vector defined as $[\nabla_{\mathbf v(-1,i)}\text{AUM}(\mathbf{\hat{y}})+\nabla_{\mathbf v(1,i)}\text{AUM}(\mathbf{\hat{y}})]/2$.
When running our gradient descent algorithm, we have observed that it empirically almost always stays at differentiable points (i.e., columns of directional derivative matrix are equal).
However, we have observed a few cases where the gradient descent algorithm visits non-differentiable points, for which there is at least one row in the directional derivative matrix with entries that are not equal. 
Examples of such directional derivative rows $[\nabla_{\mathbf v(-1,i)}\text{AUM}(\mathbf{\hat{y}}),\nabla_{\mathbf v(1,i)}\text{AUM}(\mathbf{\hat{y}})]$ that we have observed include $[-0.0019, -0.0011]$ and $[-0.0006, 0]$, both of which indicate that the loss would increase if the predicted value is decreased. 
We also perform line search via grid search in order to obtain a step size $\alpha^{(j)}$ which results in the largest decrease in AUM.
Also let $\beta^{(j)}$ be an intercept or threshold with minimal error after taking the line search step (it only affects the label error/accuracy and not the AUM).
Note that this intercept can be efficiently computed at the same time as the AUM and its directional derivative matrix, by a simple linear scan over all thresholds.
We then perform gradient descent updates for each iteration $j\in\{0,1,\dots\}$ via
\begin{equation}
    \mathbf{\hat y}^{(j+1)} = \mathbf{\hat y}^{(j)} - \alpha^{(j)} \mathbf {\bar D}_\text{AUM}(\mathbf {\hat y}^{(j)}) + \beta^{(j)}.
\end{equation}

\begin{figure*}[t]
\vskip 0.2in
\begin{center}
\includegraphics[width=\textwidth]{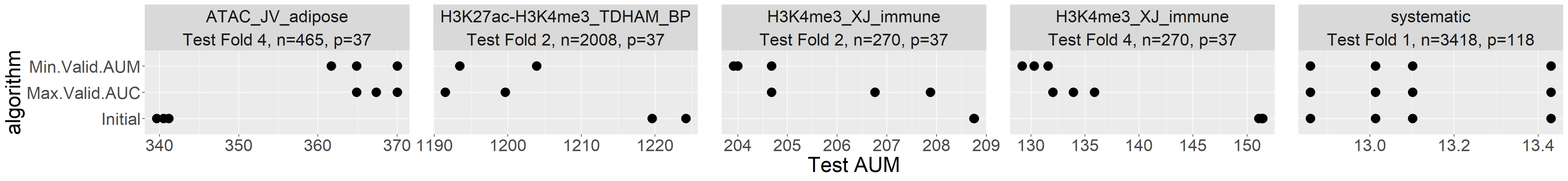}
\includegraphics[width=\textwidth]{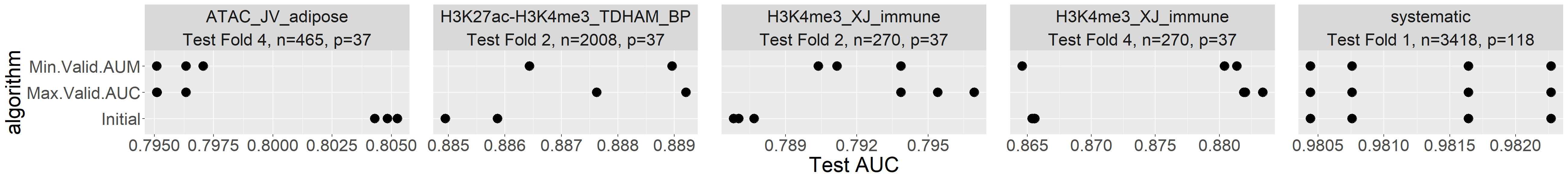}
\vskip -0.5cm
\caption{
Linear model which minimizes AUM via gradient descent results in maximizing AUC on held-out test data.
Algorithm (y axis) shows how number of iterations was chosen, by minimizing validtion AUM, maximizing validation AUC, or taking the initial iteration.
Each of five columns from left to right shows results for four random seeds on a given data set and test fold (n,p=number of examples and features for that entire data set).
}
\label{fig:test-aum-comparison}
\end{center}
\vskip -0.2in
\end{figure*}

\subsection{Gradient descent algorithm for linear model}
\label{sec:linear-model}
In the context of making predictions on a held-out test set, we consider a linear model $f(\mathbf x) = \mathbf w^\intercal \mathbf x + \beta$ parameterized by a weight vector $\mathbf w\in\mathbb R^p$ to optimize via gradient descent, and an intercept $\beta\in\mathbb R$ optimized via a linear scan to find a threshold with minimum label error.
Let $\mathbf X\in\mathbb R^{n\times p}$ be the feature/input matrix, so $\mathbf X \mathbf w+\beta\in\mathbb R^n$ is the vector of predicted values on the train set.
We consider an intialization $\mathbf{w}^{(0)}$ based on minimizing a convex squared hinge loss with L1 regularization \citep{Hocking2013icml}.
This convex loss function has a minimum for each example at predicted values that achieve minimum label errors, so we expect this initialization to have large accuracy but not necessarily large AUC.
Again let $\alpha^{(j)}$ be the line search step size, and let $\beta^{(j)}$ be an intercept with minimal train error.
We perform gradient descent for each iteration $j\in\{0,1,\dots\}$ via
\begin{equation}
    \mathbf{w}^{(j+1)} = \mathbf{w}^{(j)}   - \alpha^{(j)}\mathbf X^\intercal \mathbf {\bar D}_\text{AUM}(\mathbf X \mathbf w^{(j)}   + \beta^{(j)}).
\end{equation}
To regularize the model for the experiments in this paper, we use early stopping (number of  iterations chosen by minimizing AUM or maximizing AUC with respect to a held-out validation set).

\section{Empirical Results}
\label{sec:results}
We empirically study AUM minimization in the context of binary classification and changepoint detection problems.
Our goal is to demonstrate that AUM minimization can result in AUC maximization with respect to train and test data.
\subsection{AUM gradient descent optimizes train AUC}

In this experiment our goal was to demonstrate that minimizing the AUM results in maximizing the AUC in the train set. 
We used the chipseq data (a benchmark for labeled changepoint detection) from the UCI repository \citep{asuncion2007uci}, treating each (set.name, fold) as a different train set, with pre-processing as previously described.\footnote{
\small \href{https://github.com/tdhock/feature-learning-benchmark}{github.com/tdhock/feature-learning-benchmark}}
In brief, for each labeled example, changepoint models were computed for a range of penalty values, which resulted in models with a range of changepoints (some with few changepoints, others with many).
Then the label error rate for each model was computed, along with a penalty $\hat \lambda_i$ which resulted in minimum label errors for each labeled example $i$.
Finally these penalty values were used as the initial prediction vector $\mathbf{\hat y} = [ -\log\hat\lambda_1 \cdots -\log\hat\lambda_n]$, which was used as the optimization variable in an AUM gradient descent algorithm. 
A line search was used for the step size so that the AUM was guaranteed to decrease at each iteration. 
Overall there were 68 different train sets with the number of labeled examples ranging from $n=7$ to 1011, and the total number of breakpoints in error functions ranging from $B=47$ to 3104.

We expected that by using the predicted values directly as the optimization variable in gradient descent, we should be able to obtain ROC curves that were significantly different from the initialization (hopefully with larger AUC, even though the initialization came from minimizing the label error independently for each example).
In a small number of train sets we observed that the optimization resulted in little or no change to both AUM and AUC; this happened when the initialization was close to or at a stationary point of the AUM (for example, when AUM=0 and AUC=1).
However in most of the train sets we observed that minimizing the AUM results in increased AUC (on the train set). 

In one representative train set (H3K4me3\_XJ\_immune fold 4 which has $n=54$ labeled examples with a total of $B=347$ breakpoints in error functions), we observed that the AUC and label error rate increases during the same iterations that the AUM decreases (Figure~\ref{fig:aum-optimized}, left).
Before optimization the ROC curve was highly non-monotonic with a sharp point in the upper left corner; after AUM optimization the ROC curve became more regular with increased area (Figure~\ref{fig:aum-optimized}, middle).
This result suggests that AUM gradient descent can be used to maximize the AUC, although the label error rate also increases.

We performed a second experiment, this time optimizing  $p=27$ weights in a linear model parameter vector which was used to compute a prediction for each of the $n=54$ labeled examples.
The weights were initialized by using a gradient descent algorithm to minimize an un-regularized squared hinge loss that is a convex relaxation of the label error \citep{Hocking2013icml}.
We expected the constraints of the linear model to reduce the accuracy with respect to the previous experiment (direct optimization of predicted values).
We observed that after optimizing the weights using AUM gradient descent, the train AUC remained the same, but the train error rate increased (Figure~\ref{fig:aum-optimized}, right).
This experiment shows that the constraints of a linear model can prevent AUM minimization from resulting in AUC maximization (even in a data set for which it is possible to achieve larger AUC values).

\begin{figure*}[t]
\vskip 0.2in
\begin{center}
 \includegraphics[width=0.49\textwidth]{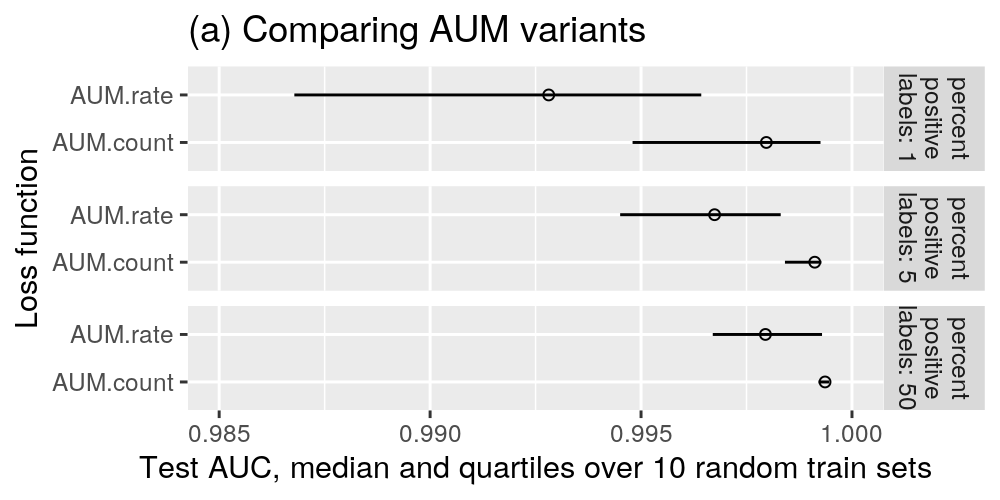}
 \includegraphics[width=0.49\textwidth]{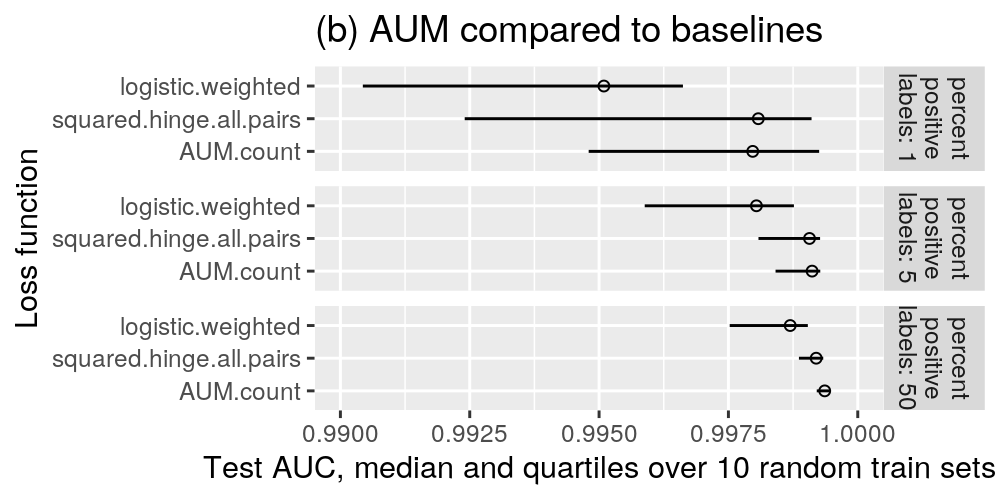}
\vskip -0.5cm
\caption{
Test AUC in binary classification (images of 0/1 digits). Fixed test set with balanced classes (50\% positive, 50\% negative labels), and three train sets with different class imbalance (1\%, 5\%, 50\% positive labels). \textbf{Left (a):} AUM count is more accurate than AUM rate. \textbf{Right (b):} AUM count is at least as accurate as baselines, and sometimes more accurate.}
\label{fig:test-binary}
\end{center}
\vskip -0.2in
\end{figure*}

\subsection{Test AUM and AUC in changepoint problems}

In these experiments, the goal was to demonstrate that test AUC correlates with test AUM using our proposed linear model based on AUM minimization.
We considered supervised changepoint detection data sets from a public repository.\footnote{\small\href{https://github.com/tdhock/neuroblastoma-data}{github.com/tdhock/neuroblastoma-data}} 
It included pre-defined fold IDs that we used to define train/test splits over labeled examples of changepoint detection problems.
In the five train sets that we analyzed (Figure~\ref{fig:test-aum-comparison}), the number of labeled examples ranged from $n=216$ to 3322, the number of features from $p=26$ to 117, and the total number of error function breakpoints from $B=1275$ to 7732.
As explained in Section~\ref{sec:linear-model}, our initialization and baseline was a linear model learned via gradient descent on a convex squared hinge loss with L1 regularization \citep{Hocking2013icml}.
To determine the extent to which the result depends on random initialization, we used four different seeds to select four different initial models. 
Some of the random seeds resulted in the same initial weights, despite having different seeds.

For each train set and random seed we used AUM gradient descent and selected the number of iterations via 4-fold cross-validation (take mean validation AUC/AUM over folds, then select iterations by maximizing AUC or minimizing AUM).
In general we observed that both methods for selecting the number of iterations resulted in increased test AUC and decreased test AUM (middle three panels of Figure~\ref{fig:test-aum-comparison}). 
However for some data the AUM gradient descent did not improve over the initialization (left and right panels).
This can be explained because some splits had very different train and test sets (left) or 
the selected number of gradient descent iterations was zero
(right). 
Overall these real data experiments indicate that AUM minimization often results in AUC maximization on held-out test data.

\begin{figure*}[t]
\vskip 0.2in
 \begin{center}
\includegraphics[width=\textwidth]{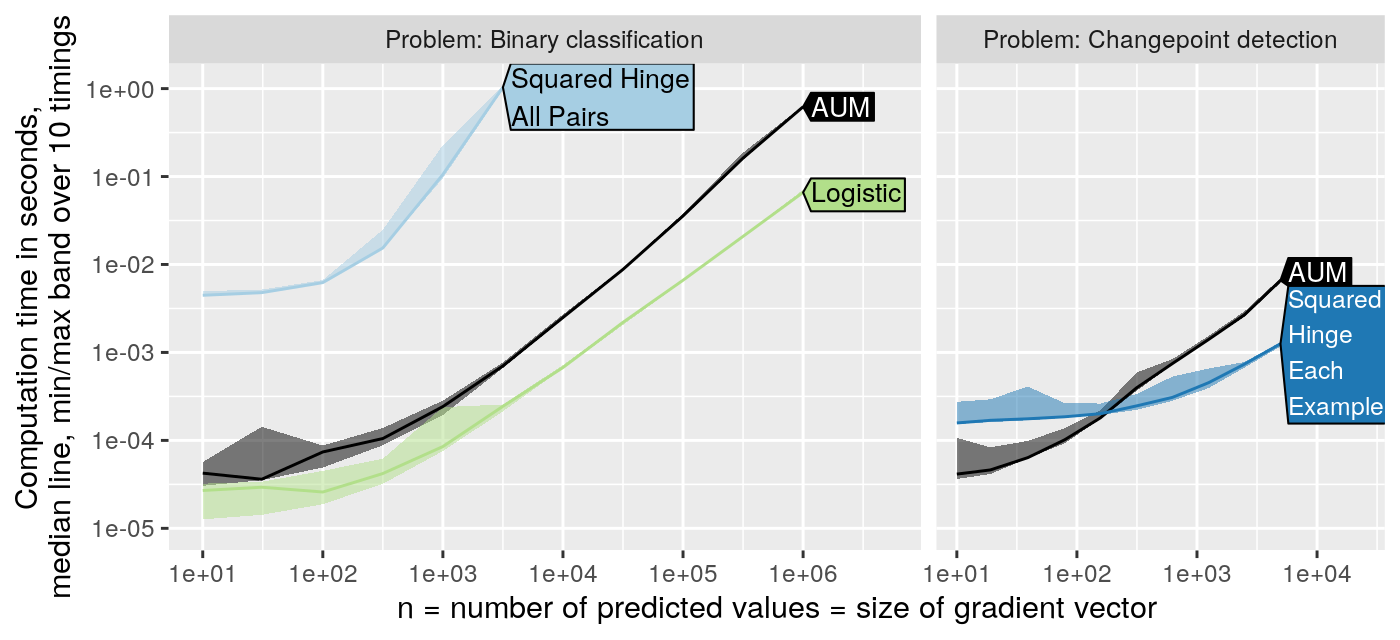}
\vskip -0.5cm
\caption{
Speed comparison, time to compute gradient of various loss functions for data sets with variable number $n$ of labeled examples. 
\textbf{Left:} for binary classification our timings suggest that AUM is $O(n\log n)$, faster than a $O(n^2)$ squared hinge loss summed over all pairs of positive/negative examples, and slower than $O(n)$ Logistic loss summed over all labeled examples.
\textbf{Right:} for changepoint detection problems we observed that AUM is asymptotically similar to, but slower than, a squared hinge loss summed over labeled examples.}
\label{fig:test-speed}
\end{center}
\vskip -0.2in
\end{figure*}

\subsection{Test AUC in binary classification}

The goal in this section was to study the accuracy of our proposed AUM loss function in unbalanced binary classification problems.
Our experiment used the zip.train and zip.test image classification data\footnote{\url{https://web.stanford.edu/~hastie/ElemStatLearn/}} from \citep{Hastie2009}.
Each input is a $16\times 16$ image which is represented by a vector $x\in\mathbb [-1,1]^{256}$.
There are 10 classes (one for each digit), but we only used two classes (0/1) in order to study binary classification.
For both the \texttt{zip.train} and \texttt{zip.test} files, we discarded some $y_i=0$ labels in order to obtain sets with equal numbers of positive and negative labels (total 2010 in train, 528 in test).
Then we generated 10 different train sets by randomly selecting 1000 examples with a class balance in $\{1\%, 5\%, 50\%\}$.
The goal is to learn from the unbalanced train set data and then provide accurate predictions (as measured by AUC) on the balanced test set.
The learning algorithms that we considered in our comparison were all based on gradient descent with constant step size and early stopping regularization. 
The step size and number of iterations hyper-parameters were chosen using grid search on a held-out validation set.
Since AUM is a full gradient method (not stochastic), we compared with several other loss functions using the full gradient method.

In our first comparison we wanted to understand if it is beneficial to normalize the AUM (use relative error rates rather than absolute error counts). 
In equations (\ref{eq:AUM}--\ref{eq:AUM-computation}) we defined the AUM as the area under the minimum of false positive and false negative \emph{counts}, but we could instead use \emph{rates}.
To compute AUM using rates, we need only change FPT/FNT functions in the minimum function (\ref{eq:M}) to FPR/FNR.
In Figure~\ref{fig:test-binary} we refer to this variant as \emph{AUM.rate}, and the original version as \emph{AUM.count}. 
It is clear from Figure~\ref{fig:test-binary}(a) that the \emph{AUM.count} loss function variant has consistently larger test AUC than \emph{AUM.rate}.
The advantage of \emph{AUM.count} becomes greater as the class imbalance is increased.
For example, with no class imbalance median test AUC for \emph{AUM.count} is 0.9993 and for \emph{AUM.rate} is 0.9979 ($p=0.03$ in one-sided $t_9$-test); for large class imbalance with 1\% positive labels median test AUC for \emph{AUM.count} is 0.9979 and for \emph{AUM.rate} is 0.9928 ($p=0.004$).
These results suggest that it is not beneficial to normalize the AUM, and it is more accurate to simply minimize the absolute min(FP,FN) counts.

In our second comparison we wanted to compare AUM optimization to standard baseline loss functions for binary classification. 
As a baseline for margin loss functions (summed over labeled examples), we used a weighted version of the logistic loss $w_i \log[ 1 + \exp(y_i f(\mathbf x_i))]$. 
The weights $w_i\in\mathbb R$ are defined in order to ensure that the sum of weights over all negative/positive examples is the same \citep{Menon2013}.
The total number of positive examples is $N_1=|\{i:y_i=1\}|$ and negative examples is $N_{-1}=|\{i:y_i=-1\}|$;
the weights are therefore $w_i=1/N_{y_i}$.
For example when the class balance is 10\%, there are $N_1=100$ positive examples and $N_{-1}=900$ negative examples. 
Each positive example $i$ has a weight of $w_i=1/100$, and each negative example $i$ has a weight of $w_i=1/900$.
The total weight over each class $y\in\{1,-1\}$ is $\sum_{i:y_i=y}w_i=1$, which makes both classes equally important in the loss function.
As a baseline for pairwise loss functions (summed over all pairs of positive $y_j=1$ and negative $y_i=-1$ examples), we used a squared hinge loss, $[1-f(\mathbf x_j)+f(\mathbf x_i)]_+^2$.
Both baselines were chosen because they are supposed to maximize AUC (rather than the accuracy rate).
We observed in Figure~\ref{fig:test-binary}(b) that the learning algorithm using the AUM loss had test AUC values consistently better than or competitive with the baselines.
The AUM loss had median test AUC that was consistently larger than the weighted logistic loss, with the largest difference for 1\% positive labels (AUM=0.9979, logistic=0.9950, $p<0.06$).
There were smaller differences in median test AUC between the AUM loss and the squared hinge all pairs loss, with the largest difference for balanced labels (AUM=0.9993, pairs=0.9991, $p<0.19$).
Overall these experiments demonstrate that the AUM loss is highly competitive with the baseline loss functions, even when the train data set has large class imbalance.

\subsection{Speed comparison}

The goal of this section is to show that AUM gradient computation has comparable speed to existing loss functions. 
Since the AUM is a full gradient algorithm (computes gradient with respect to full training set), we compared to other full gradient methods (no stochastic / minibatch methods).
In the case of binary classification, we compared the gradients of three loss functions, as shown in (Figure~\ref{fig:test-speed}, left): squared hinge all pairs $O(n^2)$, AUM $O(n\log n)$, and logistic $O(n)$.
For a fixed budget of computation time (1 second), the gradient of the AUM loss can be computed for much larger problems ($n=$ millions) than the na\" ive square hinge all pairs approach ($n=$ thousands).
Although the gradient for the AUM has a slower asymptotic runtime than logistic loss, an AUM-based model creates substantially better test data AUC, especially with imbalanced datasets.

We also did a speed test in the context of real changepoint detection problems, by comparing the AUM to a squared hinge loss summed over each of the $n$ labeled examples \citep{Hocking2013icml}. 
We observed that the AUM gradient is slower than the squared hinge loss in terms of asymptotic complexity (Figure~\ref{fig:test-speed}, right).
For example we observed about 10ms for AUM versus 1ms for the squared hinge for about $n\approx 5000$ examples with approximately 5 breakpoints in each example-specific error function.
Overall these data show that the AUM has comparable speed to previously proposed loss functions for AUC optimization in binary classification, and changepoint detection.

\section{Discussion and conclusions}
\label{sec:discussion}

In this paper we proposed the new AUM loss function which can be used in the context of prediction problems with false positive/negative rates such as in supervised binary classification and changepoint detection.
We showed that the AUM can be interpreted as an L1 relaxation of a loss function that sums the min(FP,FN) rate over all distinct points on the ROC curve.
Minimizing the AUM therefore encourages the points on the ROC curve to be in the upper left of the (FPR,TPR) space.
We proposed a new algorithm for efficiently computing the AUM and its directional derivatives.
For $n$ labeled training examples and $B$ total breakpoints in all error functions, our algorithm for computing the $n\times 2$ directional derivative matrix is $O(B\log B)$ time, which is much faster than previous $O(n^2)$ full gradient approaches that na\" ively sum over all pairs of positive and negative examples.

In our empirical comparisons with several baseline full gradient algorithms, we observed comparable or better test AUC using our proposed method.
We have shown that in binary classification, the AUM loss often out-performs baseline margin and pairwise losses (especially with un-balanced labels in the training data).
We observed that in changepoint detection problems with non-monotonic FP/FN functions, there is a tradeoff between AUC and accuracy that does not exist in binary classification problems.
If maximizing accuracy with respect to the labels is important, we can use existing convex loss functions which are surrogates to minimizing label error rates \citep{Hocking2013icml}.
If maximizing AUC is important, then we can minimize our new AUM loss function which we have shown empirically results in AUC maximization (but lower accuracy).
This tradeoff between AUC and accuracy means that the max accuracy model results in a highly non-monotonic ROC curve with many sub-optimal points, whereas the max AUC model has a more regular ROC curve (Figure~\ref{fig:aum-optimized}).
Overall our empirical results indicate that the AUM is a useful surrogate loss function for optimizing ROC curves.

For future work, we have already started exploring the connection between AUM and other AUC relaxations, in order to create new AUC optimization algorithms.
In fact, using an L1 relaxation pairwise loss (hinge with margin of zero) is quite similar to our approach (AUM integrates the min of FP and FN over all thresholds, whereas pairwise loss integrates the product).
This suggests that it should be possible to compute the full gradient for a pairwise loss in log-linear time, so we are currently exploring a new algorithm based on that idea.
Additionally, we would like to consider several variants of our new AUM loss function.
In this work we proposed a full gradient algorithm (takes a step defined by all training examples), and we would additionally like to explore a batch variant (takes a step defined by a subset of training examples).
Although that would be non-trivial since the AUM is not separable over training examples, it would be interesting to compare a batch variant with recent work on stochastic gradient methods for AUC optimization \citep{ying2016stochastic}.
Finally our current algorithm used either constant step size or a grid search, but a faster learning algorithm could potentially be obtained by exploiting the piecewise linear nature of the AUM during the step size computation.

\paragraph{Author statement.} The authors declare that they have no conflict of interest. TDH contributed to the conception of ideas, proving theorems, implementing the algorithms, making figures, writing and revising the text.
JH contributed to analysis of data, creation of figures, writing and revising the text.

\paragraph{Reproducible Research Statement.} All of the software and data required to make the figures in this paper can be downloaded from \url{https://github.com/tdhock/max-generalized-auc}. 
An R package with C/C++ code that implements Algorithm~\ref{alg:gradient-computation} is available at  \url{https://github.com/tdhock/aum}.

\bibliography{refs}
\bibliographystyle{abbrvnat}

\end{document}